# PerSum: Novel Systems for Document Summarization in Persian

Saeid Parvandeh[1], Shibamouli Lahiri[2], Fahimeh Boroumand[3]

[1]Tandy School of Computer Science, University of Tulsa, Tulsa, OK 74104, USA

[2]Computer Science and Engineering, University of Michigan, Ann Arbor, MI 48109, USA

[3]Department of Computer Engineering, Borujerd Branch, Islamic Azad University, Borujerd, Lorestan, Iran

saeid-parvandeh@utulsa.edu, lahiri@umich.edu, f.boroomand92@gmail.com

**Abstract**

*In this paper we explore the problem of document summarization in Persian language from two distinct angles. In our first approach, we modify a popular and widely cited Persian document summarization framework to see how it works on a realistic corpus of news articles. Human evaluation on generated summaries shows that graph-based methods perform better than the modified systems. We carry this intuition forward in our second approach, and probe deeper into the nature of graph-based systems by designing several summarizers based on centrality measures. Ad hoc evaluation using ROUGE score on these summarizers suggests that there is a small class of centrality measures that perform better than three strong unsupervised baselines.*

## 1. Introduction

The Persian language is spoken by 110 million people worldwide, and is considered one of the official languages in three different countries – Iran, Afghanistan, and Tajikistan.[1] Persian (also known as *Farsi*) has a rich literary and political history spanning at least two and a half millennia, and has influenced a number of modern languages including Arabic, Urdu, Hindi, Turkish, and Uzbek. Prominent Persian poets from the Middle Ages include Rumi, Hafez, Saadi Shirazi, Omar Khayyam, Ferdowsi, Nizami, Shams Tabrizi, and Attar of Nishapur. Persian literature is vast and complex, and spans the geographical boundaries of a multitude of countries, permeating and pervading in the process their unique culture and art forms, and infusing their social psyche since time immemorial with ever-fresh forms of literary tapestry.[2]

---

[1]Source: https://en.wikipedia.org/wiki/Persian_language.

Despite having such a large speaker population and a vibrant presence in many countries, Natural Language Processing in Persian has been stymied by a lack of publicly available corpora and tools, which has only very recently started to change (Seraji, 2013; Seraji, 2015). In particular, research studies in Persian document summarization are fairly new and limited, and there is still a lot of room to try different approaches (please see Section 2 for a discussion of related recent work).

In this paper, we address the problem of document summarization in Persian language from two different angles. In our first approach, we used a well-known Persian news corpus (Hamshahri (AleAhmad et al., 2009)) to evaluate Parsumist – a popular Persian document summarization framework (Sections 3 and 4). The reason we chose to re-implement Parsumist is because it is one of the most highly cited and most detailed Persian summarization systems. Also, it is simple, flexible, modular, intuitive, and relatively easy to understand and implement. Therefore, a re-implementation of Parsumist, and evaluation on a large realistic corpus would give us a unique opportunity to peer into the strengths and weaknesses of such a traditional system. Our results indicate that graph-based systems perform better than Parsumist-based systems, and the choice of threshold is of particular importance for Parsumist. In our second approach, we used centrality measures on sentence networks (similar to TextRank (Mihalcea and Tarau, 2004) and LexRank (Erkan and Radev, 2004)) to perform single and multi-document summarization on the recently released Pasokh corpus (Sections 5 and 6). Pasokh consists of human-constructed summaries of news articles, and happens to be the first publicly available research corpus of its kind in the Persian language (Behmadi Moghaddas et al., 2013). It should be noted that we are the first to bring centrality-based approaches in Persian document summarization, and we are also the first to use ROUGE (Lin, 2004) in the above setting. Our results indicate that different centrality measures perform differently in this task, and only about half of the centrality indices perform better than strong unsupervised baselines.

Our contributions include: (a) extending a popular Persian document summarization framework in multiple ways, and measuring its impact on a large, realistic corpus of news articles, (b) extensive human evaluation to identify the strengths and weaknesses of particular extensions, (c) introduction of centrality-based approaches to Persian document summarization for the first time (both single and multi-document), and (d) introduction of three intuitive baselines, and variations of the standard evaluation measure ROUGE – to build the first meaningful and valid comparison point on a publicly available annotated dataset.

---

[2]https://en.wikipedia.org/wiki/Persian_literature

This paper is organized as follows. Section 2 discusses existing work (mostly recent) in Persian document summarization. We describe Parsumist (Shamsfard et al., 2009a; Shamsfard et al., 2009b) and our extensions in Section 3, followed by human evaluation on the Hamshahri corpus in Section 4.[3] Centrality-based summarization methods are explained in detail in Section 5, followed by the results on single and multi-document summarization on Pasokh corpus in Section 6. Section 7 concludes the paper, offering future research directions, and discussing and outlining potential problems and limitations of our work. Systems, tools, and corpora are introduced and explained as and when they appear in the text for the first time.

## 2. Related Work

Arguably the first Persian text summarization system was FarsiSum (Hassel and Mazdak, 2004). It used a client-server application written in Perl, and was little more than an earlier implementation – SweSum (Dalianis, 2000) – augmented by a Persian stop word list. No evaluation was reported in Hassel and Mazdak (2004)'s study.

Zamanifar et al. (2008) reported the next study using lexical chains, word clustering, and a *summary coefficient* for each cluster. Words were clustered by their *co-occurrence degree* assessed by a bigram language model. Results on 60 Persian news articles showed that Zamanifar et al. (2008)'s approach was superior to FarsiSum (in terms of precision and recall) at compression ratios of 30%, 40%, and 50%.

Berenjkoob et al. (2009) explored the qualitative and quantitative merits of incorporating stemming and stop word removal in Persian document summarization.

Parsumist – one of the most important research efforts in Persian summarization – came in 2009 (Shamsfard et al., 2009a; Shamsfard et al., 2009b). Parsumist used an elaborate pipeline of three stages – preprocessing, analysis, and content selection. In the preprocessing stage, stop words were removed and content words were mapped to a *sense hierarchy* akin to WordNet (Miller, 1995). In the analysis stage, ten features were used to score a sentence, followed by a selection and redundancy removal step. In the final selection stage, anaphora were resolved and sentences were re-ordered to restore coherence and temporal cues. Parsumist was shown to perform better than FarsiSum (in terms of precision and recall) on documents from different genres.

---

[3]Note that we were not the first to use Hamshahri in Persian document summarization. Kamyar et al. (2011) were the first.

A flurry of research activity ensued after Parsumist. Kiyoumarsi and Esfahani (2011) introduced fuzzy logic in Persian document summarization. They reported performance superior to four other systems in a simulation study. Zamanifar and Kashefi (2011) discussed AZOM – a summarizer that takes into account the implicit or explicit structure of a document (in terms of paragraphs, blocks, etc). AZOM performed better than two state-of-the-art approaches, and a "flat summary" baseline. Tofighy et al. (2011) used a more rigorous structure-based approach, leveraging *block nesting* and sibling blocks. The performance was better than FarsiSum at 30-40% compression ratio, but the precision and recall values were lower compared to AZOM.

Bazghandi et al. (2012) clustered sentences based on the *semantic similarity* of their content words. Semantic similarity between two words was defined as a variant of their PMI (pointwise mutual information). Sentences were clustered using a particle swarm optimization (PSO) algorithm. The system achieved competitive results with traditional clustering approaches.

Shakeri et al. (2012) constructed an undirected sentence network similar to LexRank (Erkan and Radev, 2004) and TextRank (Mihalcea and Tarau, 2004). They applied the system to ten Persian scientific papers at a compression ratio of 50%. The network-based system performed significantly better than FarsiSum against a human-generated gold standard (in terms of precision, recall, F-score, and ROUGE-1).

An interesting study employing Analytical Hierarchy Process (AHP) to Persian document summarization was discussed in (Tofighy et al., 2013). The authors arranged six existing summarization features (word frequency, keywords, headline word, cue word, sentence position, and sentence length) in an AHP matrix to assess their relative importance, and to come up with an optimal selection of sentences. Results showed better F-score than FarsiSum at compression ratios of 30% and 40%.

Finally, Nia (2013) did his Master's thesis in generating gold standards for Persian document summarization using the *pyramid method* (Nenkova et al., 2007).

It should be noted from the discussion in this section that while the studies in Persian document summarization have been numerous, they are mostly recent, and as a result, not very mature. For example, few of them used the recently released Pasokh corpus of annotated summaries, and very few explored a large and realistic corpus such as Hamshahri. Furthermore, no studies looked into centrality-based (also known as *graph-based*) approaches to summarization. In this paper, we bridge these gaps.

**3. Parsumist and Its Extensions**

Document summarization systems have traditionally been classified into several categories: extractive vs. abstractive, single-document vs. multi-document, generic vs. query-focused, and neutral vs. opinionative. For the purpose of illustration, we briefly define these terms here. An extractive summarization system chooses sentences and/or sub-sentence units (e.g., phrases and clauses) from the original document(s) to build the summary; an abstractive one, on the other hand, creates its own language – usually by some form of a *natural language generation* component – to build a summary. There could be some extractive components within an abstractive summarizer. It should be noted that abstractive summarization is a much more difficult problem than extractive, and most existing research literature is on extractive summarization. Parsumist follows the extractive approach.

Summarization systems can work on single documents and/or multiple documents. Multiple-document summarization is generally harder than single-document summarization, owing to the additional complexity of inter-relationship between documents, which may involve comparison, contrast, conflicting points of view, differing opinions, redundant information, spurious information, and style mismatch – among others. Parsumist can work in both single-document and multi-document modes.

An interesting take on document-summarization is generic vs. query-focused. Generic summarizers give an overview of a document, without regard to any other source of information. Query-focused summarizers, on the other hand, present information that is relevant to a particular information need (i.e., a *query*). Parsumist can act as either a generic summarization system, or as a query-focused one.

Lastly, oftentimes a document presents opinionated content. This is especially prevalent in social media – news, blogs, forums, tweets, product reviews, comments, Facebook posts, etc. A good summarizer should therefore include some of the opinionated content to give its users an idea about contrasting points of view, and the resulting conflict, if any. Parsumist does not provide such a functionality, and is therefore categorized as a *neutral* summarizer, rather than an *opinionative* one.

The overall structure of Parsumist is shown in Figure 1. As mentioned before, Parsumist is an extractive summarizer, and thus saves much of the design complexity that naturally comes into play while building a full-blown abstractive system. Furthermore, Parsumist is flexible because it can perform both single and multi-document summarization. It follows a pipelined approach that is highly modular, extensible, and language-independent save the resources (Figure 1). There are four main modules – preprocessing, analysis and scoring, selection and redundancy checking, and smoothing the summary for coherence. The authors mostly focused on the first three modules, and mentioned the fourth module rather cursorily. While re-implementing Parsumist, we therefore needed to exercise a fair bit of creative license on our own. We maintained the form and spirit of Parsumist, while altering its content as and when necessary. One example is the resources used: Parsumist used a stop word list, a cue word list, and a list of Persian synsets (the authors came up with their own version of synsets). While we did have access to a stop word list, we did not have a publicly available cue word list or a list of synsets. Thus, we had to resort to alternative resources that will be described as they are introduced.

The remainder of this section is organized as follows. In Subsections 3.1-3.4, we describe the four components of Parsumist, along with details on our re-implementation of the same. In the end, we designed four systems based on Parsumist. Subsection 3.5 gives details on the multi-document summarization part of Parsumist, and Subsection 3.6 outlines Parsumist's evaluation strategy – and how we re-implemented it.

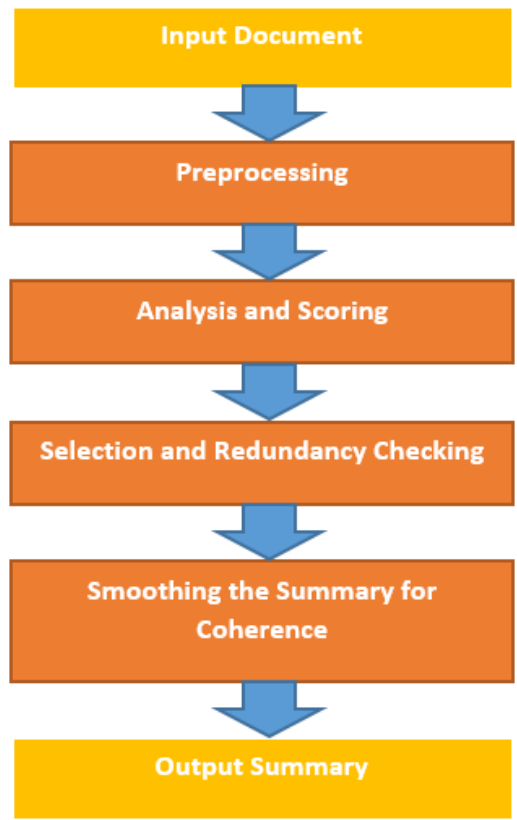

Figure 1: Pipeline structure of Parsumist

### 3.1. Preprocessing

Parsumist uses a combination of statistical, semantic, and heuristic methods. The preprocessing step consists of tokenization, stop word removal, and conceptual mapping (i.e., mapping words and phrases to synsets). Note that tokenization of free-form Persian text is rather non-trivial because of encoding problems, mixed ordering (i.e., presence of intermingled right-to-left and left-to-right sequences of characters), omitting Ezafe in Ezafe construction, and irregularities in word segmentation. While Parsumist uses a home-grown tokenizer, we used the tokenizer made available by Mojgan Seraji as part of the Persian language processing toolkit she developed (Halácsy et al., 2007; Seraji, 2013; Seraji, 2015). The version we used is based on a Ruby-on-Rails framework for basic preprocessing, and a couple of Perl scripts for sentence segmentation and subsequent word segmentation. Note further that we used Hamshahri (AleAhmad et al., 2009) as our input corpus for this part, and Hamshahri is released in NCR (*numeric character encoding*) encoded XML format, so we needed to convert NCR into UTF-8 Unicode first. We used the *ltchinese* Python library[4] for this purpose.

Stop word removal requires a standard stop word list. Since the stop word list used by Parsumist authors was not publicly available, we used the stop word list that comes with the Hamshahri corpus. Similarly, Parsumist authors mapped each word and phrase to a *concept* (i.e., a synset) in a small concept hierarchy akin to WordNet (Miller, 1995). This concept hierarchy was later used to construct an undirected sentence network (please see Subsection 3.2). Since the concept hierarchy was not publicly available, we constructed our sentence networks using Latent Semantic Analysis (LSA; see Subsection 3.2 for details).[5]

### 3.2. Analysis and Scoring

As mentioned before, Parsumist uses an extractive summarization strategy for the most part. This strategy requires scoring sentences for their importance (*salience*), and subsequently selecting the highest-scored sentences that are presumably also the most important. There are many different ways to score sentences; Parsumist uses a very simple additive strategy based on surface, syntactic, and semantic features. The overall scoring function is as follows:

$$W(s_i) = \sum_j c_j p_{ij} \tag{1}$$

---

[4] https://pypi.python.org/pypi/ltchinese
[5] Note that we are not the first to introduce LSA in Persian document summarization. Poormasoomi et al. (2011) were the first.

where $W(s_i)$ is the score of the $i$-th sentence, $c_j$ is the weight of feature $j$ (assumed to be 1 by the authors), and $p_{ij}$ is the value of feature $j$ in sentence $i$. The features are as follows:

1. **Number of main words, title words, and query words.** Presence of important words in a sentence (esp. words that appeared in the document title) often indicates that the sentence is important, and should probably be included in the summary. Furthermore, note that Parsumist can work as a *query-focused* summarizer by enhancing the importance of sentences that contain query words. In our implementation, we only considered *title words* (i.e., words that appeared in the document title), and disregarded *main words* and *query words*. Thus, our summarizer is *generic* (as opposed to *query-focused*). Moreover, we used the frequency of title words in a sentence (normalized by sentence length) as the feature value.
2. **Length of the sentence.** Authors posited that shorter sentences were more likely to appear in a summary. While the merit of this decision is debatable (longer sentences presumably contain more information, and should therefore be given a higher weight), in our implementation we simply added a feature – reciprocal of the sentence length – to simulate the correct behavior.
3. **Proper nouns.** Authors correctly observed that the importance of proper nouns (*named entities* – more generally) depends on the type of document. If we are summarizing news articles, academic papers, or short stories, proper nouns usually are important because they signal the presence of a named entity. For informal media like blogs, emails, tweets and online forums on the other hand, proper nouns may be much less important. In our implementation, we used the frequency of proper nouns in a sentence (normalized by sentence length) as the feature value.
4. **English words and phrases.** The authors made an interesting observation about English words in Persian text; presence of English words usually indicates the *first appearance* of a new term in a document. Such terms are especially relevant if the document is talking about a scientific study (e.g., research articles). Our implementation uses the frequency of English words in a sentence (normalized by sentence length) as the feature value.
5. **Quotation marks.** Quotation marks in a sentence should be given due importance because they signify an important previous commentary. We used the frequency of quotation marks in a sentence (normalized by sentence length) as the feature value.
6. **Pronouns.** Most pronouns indicate some form of anaphora. Authors correctly

pointed out that we should select the sentence that contained the referent, in addition to selecting the sentence with the pronoun. However, this would require anaphora resolution, which is a hard problem for Persian. The authors therefore opted for an easier solution – reduce the score of a sentence that contains a pronoun, but do not delete it. We chose to use the same solution – with a twist. We used the frequency of *non-pronoun words* in a sentence (normalized by sentence length) as the feature value.

7. **Percentage sign ("%").** The percentage sign is used to represent results (esp. improvement) in scientific documents, and profit margins and general increments-decrements in financial documents. Hence derives its importance. We used the frequency of percentage signs in a sentence (normalized by sentence length) as the feature value.

8. **Parenthetical or descriptive sentences.** Embedded sentences and phrases – such as parentheticals – often do not add much value to the information content of a sentence, and hence can safely be deleted from a summary. This is a form of *abstractive summarization* that Parsumist designers performed. In our implementation, we chose to simply include *parentheses* as features. We used the frequency of *non-parenthesis tokens* in a sentence (normalized by sentence length) as the feature value. This was done to ensure that sentences containing parentheses are given reduced importance. Note that the original implementation probably required a dependency parser to obtain this feature. We chose to sidestep dependency parsing of Persian text, because as the tools are not very mature yet, it may yield poor results.

9. **Referential phrases.** Authors maintained that phrases that refer to other parts of a document ("in the last section", "in the previous figure", etc) should not appear unaltered in the summary. They should be changed, and/or gain lower scores. While it is true that such phrases do not contribute (much) to the information content of a sentence, we are of the opinion that they do contribute to the *style* – esp. *coherence* of the document. Leaving them out in the final summary may therefore hurt coherence. In our implementation, we chose not to include this feature, as it may require anaphora resolution and/or discourse parsing – for which tools are not that mature in Persian.

10. **Punctuation marks.** Parsumist authors were not very clear about this feature, so we chose not to include it in our implementation.

In the end, we had eight distinct features, as opposed to ten in the original study. Note that in almost all the above cases, we needed to exercise our own discretion to come up with an appropriate feature value. This was due to under-specification and lack of clarity on the part of Parsumist authors, who did not adequately specify and/or quantify most of the above features. We therefore needed to come up with our own best (educated) guesses as to what they could have been. We also had to alter parameters and introduce new ones as appropriate. For example, we came up with sentence length normalization to combat the variability of sentence length. Many of the features needed a part-of-speech tagger; we used Mojgan Seraji's tagger (Halácsy et al., 2007; Seraji, 2015) for this purpose.[6]

Once the score of each sentence is known, Parsumist constructs an undirected graph between sentences – each node is a sentence, and two nodes are connected if their similarity exceeds a certain threshold. Similarity between two nodes (sentences) is computed using lexical chains and a *synset hierarchy* akin to WordNet (Miller, 1995). The authors are very vague in describing how they computed this similarity metric. They do mention that the similarity is computed using the number of common or related words due to lexical chains and synsets, but no further details are provided. Moreover, the authors came up with their own synset hierarchy, and never released it to the public. Hence, we had to come up with our own similarity measure (instead of using the original method).

We defined the similarity between two sentences as the cosine between their LSA vectors.[7] Note that LSA relies on Singular Value Decomposition, a matrix factorization technique that comes from Linear Algebra. The basic premise of LSA is very simple; if two words have similar meaning (say "rat" and "mouse"), then they should occur in similar contexts. Hence, if we were to construct a *co-occurrence matrix* between words, and somehow compress this matrix into a low-dimensional form (known as *low-rank approximation*), then the resulting matrix will yield vectorial representations for semantically similar words that are "close" to each other in terms of cosine similarity – even if the original vectors were not close together.

The particular way LSA works is as follows: we first construct a term-document matrix, where rows are words, and columns are documents. Each word is represented as a row vector, and each document is a column vector. The particular value of a single matrix cell could be term frequency, tfidf, or even binary presence/absence (see, e.g., Chung and Pennebaker (2008)). For our implementation, we chose simple term frequency as the value.

Once the term-document matrix (say, **X**) is constructed, we perform a Singular Value Decomposition (SVD) as follows:

---

[6]Available at http://stp.lingfil.uu.se/~mojgan/tagper.html.
[7]Latent Semantic Analysis (Landauer and Dumais, 2008).

$$X = U\Sigma V^T \qquad (2)$$

where **U** and **V** are *orthogonal matrices*, and **Σ** is a *diagonal matrix*. The diagonal entries $\sigma_i$ of **Σ** are known as *singular values*. Now, say **X** was an M-by-N matrix; then **U** is an M-by-M matrix, **V** is an N-by-N matrix, and **Σ** is an M-by-N matrix.[8]

With that, we are now ready to describe the low-rank approximation (*compression*) step of LSA. Essentially, we perform a *truncated SVD*:

$$X_k = U_k \Sigma_k V_k^T \qquad (3)$$

where **X_k** is an M-by-N low-rank approximation of the original term-document matrix **X**, **U_k** is an M-by-k matrix (the first *k* columns of **U**), **Σ_k** is a k-by-k square diagonal matrix (the first *k* rows and first *k* columns of **Σ**), and **V_k** is an N-by-k matrix (the first *k* columns of **V**). As it turns out, this truncated SVD representation is a form of lossy compression that is the "closest" to the original term-document matrix **X**, in the sense that it minimizes the *Frobenius Norm* of the difference between **X** and **X_k**. **X_k** is known as the "rank-*k* approximation" of **X**, since we took the largest *k* singular values, and the leftmost *k* columns of **U** and **V** to construct **X_k**. What comes out as the most important upshot of this decomposition procedure is a fact we alluded to in the above discussion: words that are *semantically close* to each other (say "rat" and "mouse"), become closer in their *vectorial representation* (i.e., on the vector space). In other words, the original *sparse matrix* **X** becomes a *dense matrix* **X_k**, where semantically similar terms occupy similar regions of the vector space.

Finally, we project a sentence onto this dense LSA space **X_k**, as follows. We first construct a *term-vector* for the sentence (say, **s**), whose dimensionality is M-by-1. The individual elements of this vector are simple term frequencies (as before). Once this vector is constructed, we use the following equation to turn it into a compressed (low-ranked) vector **s_k** of dimensionality *k*-by-1:

$$s_k = \Sigma_k^{-1} U_k^T s \qquad (4)$$

---

[8]M is the number of unique words. N is the number of documents.

This $s_k$ is what we will henceforth refer to as the "LSA vector" of a sentence. Note that the reason we needed to perform LSA on sentences is because Parsumist uses a *semantic similarity* between sentences, and LSA yields a very good proxy for semantic similarity. It is one of the cornerstones of modern distributional semantics (see, e.g., Bruni et al. (2014)), and is reasonably free from statistical assumptions.[9] Parsumist authors used lexical chains and synsets to come up with semantic associations; we believe LSA is a finer-grained solution than that. We performed LSA on the full Hamshahri corpus, and chose the value of *k* to be 200 after initial parameter tuning.

### 3.3. Selection and Redundancy Checking

As we mentioned before, sentence selection is a key step in any extractive summarization system. Redundancy checking, however, is more germane to multi-document summarization than single-document summarization. In single-document summarization, the sentences usually describe *different* aspects of an underlying issue. In multi-document summarization, there is no such guarantee. Two sentences from two different documents may describe exactly the *same* aspect of the same issue. Since Parsumist has the option of working as a multi-document summarizer, checking and managing redundancy becomes a paramount objective.

The way Parsumist handles redundancy consists of a three-step process that is merged with sentence selection:

a) Begin with an empty summary.
b) As long as the summary length is shorter than desired, choose the sentence with highest score and minimum resemblance with the already-selected sentences.
c) Continue until the desired summary length is reached.

As we can see, the above description leaves several questions un-answered (e.g., what is the minimum resemblance threshold? How many of the previously selected sentences to consider? How to choose among previously selected sentences?). We conducted extensive parameter tuning, and came up with the following algorithm (described in Python pseudocode):

---

[9]With concomitant limitations, however; the discussion of which is outside the scope of this paper.

```
sorted_list = sort(sentences) by score in descending order
selected_sentences = []
for sentence in sorted_list:
    if len(selected_sentences) == 10: break
    if sentence_length is outside 1st and 3rd quartiles: continue
    if min_cosine_sim(sentence, selected_sentences) > threshold:
        continue
    append(selected_sentences, sentence)
```

What this algorithm does, is as follows. It keeps selecting highest-scored sentences for inclusion in the summary until the desired summary length (10 sentences) is reached. The algorithm makes sure that too long and too short sentences are excluded, by only taking into account sentences whose length falls within the 1st and 3rd quartiles of all Hamshahri sentences. It also excludes sentences that are too similar to already selected sentences, by enforcing a threshold on the *minimum cosine similarity* with the previously selected ones. Note that the 10 sentences length and minimum cosine similarity were selected as part of preliminary parameter tuning. Later, we observed that cosine thresholds of 0.1 and 0.2 produced reasonably good summaries. Among median, maximum, and minimum thresholds, minimum yielded the best results. This gives us two systems:

1. Top 10 sentences, modified Parsumist, minimum cosine threshold of 0.1
2. Top 10 sentences, modified Parsumist, minimum cosine threshold of 0.2

Next we went ahead and experimented with two more systems that are *graph-based* (also called *centrality-based*). Graph-based summarizers are one of the mainstays in extractive summarization, and we will discuss them in detail in Section 5. Briefly, they construct a network (graph) where each node is a sentence, and an edge appears between two sentences if their similarity exceeds a certain threshold. For our implementation, we chose to use the *complete graph* of sentences, and *weighted edges* with cosine similarity on LSA vectors. Then we followed the popular TextRank (Mihalcea and Tarau, 2004) and LexRank (Erkan and Radev, 2004) formalism that prescribed running a random walk on this sentence network, and scoring nodes by their PageRank value (Page et al., 1998).[10] Following recent studies, we also experimented with the *weighted degree* (or *strength*) of a

---

[10]More discussion appears in Section 5.

sentence instead of PageRank.[11] This gives rise to our last two systems in this section:

3. Top 10 sentences, PageRank on LSA vector cosine similarity graph
4. Top 10 sentences, weighted degree (strength) on LSA vector cosine similarity graph

The reason we chose to use two graph-based systems at this point was to see how (modified) Parsumist competes against these popular and powerful frameworks – on the same kind of sentence vectors. If Parsumist works better than graph-based systems, then we have a strong contender for state-of-the-art extractive summarization in Persian language. On the other hand, if graph-based systems perform better, then we have yet one more strand of evidence in support of their power, pervasiveness, ubiquity, and applicability. Our results show that graph-based systems do in fact perform better than Parsumist-based systems (cf. Section 4).

**3.4. Smoothing the Summary for Coherence**

Coherence is a thorny issue in extractive summarizers, esp. multi-document ones. In single-document summarization, different sentences present different aspects of an underlying issue. Hence, presenting extracted sentences in temporal order largely resolves the coherence problem. However, for multi-document summarization, since two sentences may refer to the same aspect of the same issue at the same time point, it becomes difficult to maintain logical coherence – since two successive sentences may talk about the same issue from two completely opposite angles.

Note that *abstractive summarizers* by default suffer *less* from the coherence problem, because the summary can be smoothed in different ways to give it a more polished and coherent look. Parsumist, however, is an extractive summarizer for the most part, so addressing coherence becomes a key issue. The way Parsumist handles coherence is by penalizing sentences containing anaphora and/or taboo words. The taboo word list can be specified by the user, which is an important flexibility.[12] Anaphoric sentences are one of the principal contributors to the coherence problem, because they do not – by definition – refer to the sources of anaphora.

Parsumist employs a few other heuristics to manage the lack of coherence in generated summaries. For bulleted lists, it chooses to either ignore the list, or select at least one sentence from each bullet. It also curbs redundancy in both single and multi-document modes at the sentence, morphological, and word-semantic levels – as described before. Sentences deemed too similar are collapsed into a single cluster, and only one representative from each cluster is chosen. Furthermore, Parsumist strives to retain the temporal ordering of sentences as much as possible – without violating the "Selection and Redundancy Removal" algorithm.

---

[11]Please see Boudin (2013) and Lahiri et al. (2014) for evidence that non-PageRank centrality measures often perform as well as or better than PageRank, when it comes to keyphrase extraction. We suspect that similar observations will hold true in the summarization domain.

[12]Parsumist authors came up with their own taboo words list – which was never made public.

In our implementation, we chose to skip coherence handling and management – for two reasons: (a) details of implementation were unclear in the Parsumist paper, and (b) existing publicly available Persian tools and resources were inadequate to handle issues like anaphora and taboo words.

### 3.5. Multi-Document Summarization by Parsumist

Existing literature on multi-document summarization prescribes two distinct ways of approaching this problem:

1. Concatenate all documents, and then run the single-document summarizer.
2. Generate single-document summaries for all documents, concatenate the summaries, and then run the single-document summarizer (esp. the redundancy eliminator) on this concatenated document to produce a more compact summary.

Parsumist authors found that there was no significant difference in performance between the two approaches. It should be noted, however, that the second approach has the obvious advantage that the redundancy eliminator is run twice, thereby (potentially) resulting in a more compact, coherent, and readable summary. The trade-off, however, betrays itself in time complexity. The second approach is more time-consuming, because the whole summarization process needs to run twice.

In our implementation, since we performed redundancy elimination implicitly rather than explicitly, and since we did not have an explicit coherence management module, we opted for the first approach – concatenate all documents, and then run the single-document summarizer once.

### 3.6. Evaluation of Parsumist

Evaluation turns out to be the Achilles' heel for most summarization systems, simply because there is no agreed-upon or universal standard of summarization. Human summarizers tend to vary a lot regarding content selection and style, and most human summarizers follow an abstractive approach to summarization, as opposed to the extractive approach followed by computational systems. Evaluation is further complicated by the fact that it is often unclear (esp. for long documents or multiple documents) which units of information are important enough to be included in the summary. This concerns both the granularity of linguistic units (Words? Phrases? Clauses? Sentences? Paragraphs?), as well as their *salience* (i.e., importance).

Extractive summarizers deal with the granularity issue by (mostly) focusing on sentences. The salience issue is resolved by adopting a scoring function (also variously known as an *objective function*) – perhaps in addition to other constraints and/or heuristics – and thus imposing a ranking (i.e., an *order*) on the otherwise unordered set of linguistic units. Evaluation is usually performed by comparing system-generated summaries (also known as "peer summaries") with *multiple* human-generated summaries (also known as "model summaries") to reduce variability. As one can deduce from this discussion, agreement among human judges tends to be fairly low for extractive summarization.

Parsumist authors created a new gold standard to evaluate their system. The gold standard consisted of documents from different domains and genres – short news articles of a few sentences to short stories comprising a few hundred sentences. No further details were given, and nor was the gold standard publicly released. The authors enlisted 20 students from Computer Engineering as human judges, and obtained at least six human (extractive) summaries for each document. Sentences were ranked for importance by the annotators; this ranking accommodates different compression ratios. Note in particular that the gold standard was *single-document*, so special adjustments and accommodations needed to be made for evaluating the multi-document mode of Parsumist. Parsumist was compared against – and found to perform better than – two then state-of-the-art baselines: FarsiSum (Hassel and Mazdak, 2004), and Karimi and Shamsfard's summarizer.

Since we did not have access to the annotated gold standard data Parsumist authors generated, we instead used the Hamshahri corpus of news articles (AleAhmad et al., 2009), and evaluated our four systems (cf. Section 3.3) *post hoc*, i.e., after the summaries have been generated. This way, we can evaluate which systems performed the best in the opinion of human judges. Our judges were three native speakers of Persian, all male, and all in their twenties, holding baccalaureate degrees in Civil Engineering, Industrial Engineering, and Management, respectively. Our judges were knowledgeable about the subject matter of Hamshahri news articles.

## 4. Results from Modified Parsumist and Graph-Based Systems

In this section, we describe the human evaluation results of our four systems (cf. Section

3.3). The systems were evaluated on 50 *topics* of Hamshahri corpus. Each topic in Hamshahri consists of several Persian news articles. Hamshahri was built for Cross-Lingual Information Retrieval (CLIR), which is a different task than summarization. We adapted the dataset to our purpose. Adapting a CLIR dataset for summarization is a challenging task, and the trade-offs are not well-understood. We are the first to experiment with such an idea in Persian, and we will show (later in this section) that Hamshahri, in fact, is very well-suited for the task of summarization.

We treated the problem as *multi-document summarization*, and concatenated all documents within a single topic as a big *topic-document*. This way, we ended up with 50 topic-documents, and ran our summarization systems on them. Parsumist, as already noted, can perform multi-document summarization in this way; so can graph-based systems like TextRank (Mihalcea and Tarau, 2004) and LexRank (Erkan and Radev, 2004). We posit that our fourth system based on weighted degree (*strength*) should be able to do the same.

We gave our human judges the title and description of each of the 50 topics, to facilitate understanding and acclimatization. Then we gave them four summaries for each topic (totaling 200 summaries), and instructed them to read the summaries very carefully. Once they have read the summaries of a particular topic, they were asked to assign a score between 1 and 7 as to how good the summaries were, where "goodness" is measured by relevance to the topic title and description,[13] and overall presentation. The scale is as follows:

1. Very bad summary
2. Bad summary
3. Somewhat bad summary
4. Borderline summary (not too good, not too bad)
5. Somewhat good summary
6. Good summary
7. Very good summary

This way, each summary of each topic was given a score by each human annotator. Note that the above scheme corresponds to a Likert Scale style annotation (Likert, 1932), and is widely used in summarization research. It allows for some amount of variability in human judgment, rather than forcing the judges to make a binary or ternary judgment. This turns out to be an important decision, because as we will see later in this section, on average the judges converge toward the middle of the scale – a phenomenon known in the social sciences as the *central tendency bias*. With strict binary or ternary judgment, the presence of such bias is eliminated, which could be both good (because now we have 2-3 crisp classes of summaries) and bad (because now the bias is removed rather artificially, and the resulting 2-3 classes may contain a jumble of different types of good/bad summaries). We are of the opinion that a nuanced judgment should be the way to go – rather than a binary or ternary decision.

---

[13]Note that it is in general impossible for the human judges to read *all* documents of all topics, and then assign a score to a summary. As a proxy, we asked them to merely read the topic title and description.

The rest of this section is organized around several research questions we asked of the data obtained from our human judges. The questions probe several topics, ranging from inter-annotator agreement and intra-annotator bias to inter-system difference and inter-topic difference.

**4.1. Question 1: How Much do the Judges Agree?**

Inter-annotator agreement is very important for all summarization systems, not only because it gives us an idea of how *hard* the task is (lower agreement implies greater difficulty), but also because it gives an upper bound (*roofline*) on the performance of a proposed computational system.

| **Judge** | **Judge 1** | **Judge 2** | **Judge 3** |
|---|---|---|---|
| **Judge 1** | 1.0 | 0.94 | 0.91 |
| **Judge 2** | 0.94 | 1.0 | 0.91 |
| **Judge 3** | 0.91 | 0.91 | 1.0 |

Table 1: Cosine similarity between judges

We represented each human judge as a vector of 200 elements (50 topics, 4 summaries per topic), and measured agreement between those vectors. As shown in Table 1, the three judges indeed have a very high cosine similarity among themselves. This is reassuring, but not completely satisfying, because the *correlation* among the judges is fairly low (cf. Tables 2-4), and in fact statistically indistinguishable from zero in at least one of the cases (between Judge 2 and Judge 3).

This shows that the task of multi-document summarization in Persian is in fact very difficult, and we should expect the performance of existing summarizers to be low.

| Judge   | Judge 1 | Judge 2 | Judge 3 |
|---------|---------|---------|---------|
| Judge 1 | 1.0     | 0.35    | 0.14    |
| Judge 2 | 0.35    | 1.0     | 0.12*   |
| Judge 3 | 0.14    | 0.12*   | 1.0     |

Table 2: Pearson correlation between judges; "*" indicates statistical indistinguishability from zero at 95% confidence level.

| Judge   | Judge 1 | Judge 2 | Judge 3 |
|---------|---------|---------|---------|
| Judge 1 | 1.0     | 0.30    | 0.13*   |
| Judge 2 | 0.30    | 1.0     | 0.05*   |
| Judge 3 | 0.13*   | 0.05*   | 1.0     |

Table 3: Spearman correlation between judges; "*" indicates statistical indistinguishability from zero at 95% confidence level.

| Judge   | Judge 1 | Judge 2 | Judge 3 |
|---------|---------|---------|---------|
| Judge 1 | 1.0     | 0.23    | 0.10*   |
| Judge 2 | 0.23    | 1.0     | 0.04*   |
| Judge 3 | 0.10*   | 0.04*   | 1.0     |

Table 4: Kendall correlation between judges; "*" indicates statistical indistinguishability from zero at 95% confidence level.

### 4.2. Question 2: Are the Judges Biased? If yes, by How Much?

It is instructive to look into *intra-annotator bias*, and see how much individual judges tilt towards a certain direction of the Likert Scale. Also, it allows us to identify relatively lenient judges and relatively harsh judges, and then adjust our evaluation accordingly.

| Judge   | Mean | Standard Deviation |
|---------|------|--------------------|
| Judge 1 | 4.61 | 1.47               |
| Judge 2 | 4.84 | 1.53               |
| Judge 3 | 4.51 | 1.62               |

Table 5: Mean and standard deviation of each of the three judges

We represented each judge as a vector of 200 elements (as before), and computed the means and standard deviations of those vectors. Table 5 shows the results. Note that Judge 2 is the most "lenient", in the sense that he gave the highest ratings on average. Judge 3 was the most "harsh" from that perspective. Compared between themselves, judges were actually quite similar in terms of their average ratings, which may point to the presence of a *central tendency bias* (more discussion below).

Considering *variability*, Judge 3's ratings are the most spread out, indicating the fact that he had the most varied set of opinions. On the other hand, Judge 1 has the lowest standard deviation, which implies that his opinion is much more concentrated and focused. Overall, all three annotators exhibit central tendency bias, because all three means are very close to the central point of the Likert Scale (which is 4), and far from the two extremities (1 and 7, respectively). This may indicate two things – not necessarily mutually exclusive: (a) the task was difficult, and annotators resorted to a middling scoring tactic, and (b) the annotators were unable to judge the goodness/badness of the summaries. We opine that the first of these alternatives is true, because all three annotators were background-checked and had no reason not to understand simple newswire text in their native language. So the upshot of this section is that: (a) yes, the annotators were biased, but not too much when compared between themselves; (b) there was a clear central tendency bias; and (c) the central tendency bias was due to the difficulty of the task, and not due to annotator incompetence or laziness.

**4.3. Question 3: Are the Judges Similar?**

We already observed that owing to central tendency bias, mean ratings of all three judges were quite close to each other. We performed *repeated measures ANOVA* (also called *paired ANOVA* (equivalent to paired t-test on multiple samples)) to find out if the mean ratings are statistically significantly different.[14] The results show that they are in fact not significantly different from each other (i.e., we failed to reject the null hypothesis) at 95% confidence level. That is, judges are similar to each other in terms of mean ratings. This establishes beyond doubt the presence of central tendency bias as a ubiquitous phenomenon in our Likert Scale style annotation.

**4.4. Question 4: Are the Four Systems Different from Each Other? How Different Are They?**

We would like to know how our systems performed on Hamshahri as multi-document summarizers. Recall that we have four systems (cf. Section 3.3) – two based on Parsumist, and two graph-based. How did they perform compared to each other? To answer this question, we first needed to do *annotator standardization*.

---

[14]The p-value was computed from this website: http://graphpad.com/quickcalcs/PValue1.cfm

Recall from Section 4.2 that our annotators were biased – they had different mean overall ratings, and different standard deviations. To remove this *intra-annotator bias*, we computed a *z*-score for each annotator:

$$z = \frac{x - \mu}{\sigma} \qquad (5)$$

where *x* is a particular rating, $\mu$ is the mean of all ratings assigned by one annotator, and $\sigma$ is the standard deviation of the annotator. The *z*-score effectively tells us how many standard deviations away from the mean a particular rating lies.

| System | Mean Rating |
|---|---|
| **System 1** | -0.155 |
| **System 2** | -0.069 |
| **System 3** | 0.140 |
| **System 4** | 0.084 |

Table 6: Mean ratings of four summarization systems

Each annotator was represented as a vector of 200 elements, and each element was standardized by Equation (5), thus yielding a 200-length *z*-score vector. Next, we represented a summarization system as a vector of 50 elements, where each element corresponds to a Hamshahri topic (Hamshahri had 50 topics), and the element value is the mean of three *z*-scores from three annotators. Then we performed paired ANOVA (like Section 4.3) on the four vectors (corresponding to four summarization systems). Results indicate that there is – in fact – some statistically significant difference among the mean ratings of the four systems at 95% confidence level. Mean ratings of the four systems are shown in Table 6. Note that the ratings could be positive or negative, because they are averaged across multiple *z*-scores, and *z*-scores can be positive or negative. Note also that System 3 (PageRank on sentence network) has the highest mean rating, followed by System 4 (weighted degree on sentence network), System 2 (modified Parsumist), and System 1 (modified Parsumist), respectively. This shows that graph-based systems are undoubtedly better than the systems we implemented by modifying Parsumist. It also shows that the choice of minimum cosine threshold is important – in particular, System 2 performs better than System 1, and the only difference between the two systems is in cosine threshold.

The next question that arises is: is any of the systems statistically significantly better than any other system? To answer this question, we performed *paired t-tests* among all system pairs (with Bonferroni Correction for multiple comparison (adjusted α = 0.05/6)). Results indicate that indeed, Systems 1 and 3, Systems 1 and 4, and Systems 2 and 3 are statistically significantly different from each other at 95% confidence level (by one-tailed p-value). According to magnitude of the p-value, Systems 1 and 3 are maximally different from each other (lowest p-value), followed by Systems 1 and 4, and then by Systems 2 and 3. What this implies is that Systems 3 and 4 (the graph-based systems) are significantly better than System 1 (modified Parsumist), and System 3 (graph-based) is significantly better than System 2 (Parsumist-based). Hence, graph-based systems are better on Persian extractive summarization than Parsumist-based systems.

**4.5. Question 5: Are Similar Systems Similar in Terms of Performance?**
Now consider two "system clusters": one on graph-based systems, and another on Parsumist-based systems. Are the two graph-based systems and the two Parsumist-based systems more similar *within* themselves than *between* each other? In other words, are the clusters more tightly knit within themselves than between each other? To see this, we measured cosine similarity between all pairs of system vectors (each of length 50). Table 7 shows the results.

| System Pair | Cosine Similarity |
|---|---|
| **Systems 1 and 2** | 0.61 |
| **Systems 1 and 3** | 0.34 |
| **Systems 1 and 4** | 0.38 |
| **Systems 2 and 3** | 0.30 |
| **Systems 2 and 4** | 0.30 |
| **Systems 3 and 4** | 0.91 |

Table 7: Cosine similarity between system pairs

As seen from Table 7, the highest cosine similarities occur *within* clusters: graph-based systems (Systems 3 and 4) achieve a cosine of 91% between themselves, whereas Parsumist-based systems (Systems 1 and 2) achieve a cosine of 61% between themselves. All other pairs are *between-group pairs*, and they achieve much lower cosines.

To probe this phenomenon further, we took the *vector average* of Systems 1 and 2, and Systems 3 and 4, to come up with two *prototype vectors* – one for each of the "system clusters". These two prototype vectors achieve a cosine of 38% between themselves, which is indeed much lower than within-cluster cosines.

What this discussion shows is that *similar systems are similar* in terms of human-assigned ratings, and dissimilar systems are dissimilar.

**4.6. Question 6: How Are the 50 Topics Different in Terms of Summarization?**

Now that we have analyzed and dissected the annotators and the four systems, the next question to ask is: how are the 50 Hamshahri topics different in terms of summarization? Are they all equally easy/hard to summarize? Or is there some difference? Note that the answer to these questions has important implications in designing future Persian summarizers. For example, if we can show that some topics are inherently more difficult to summarize than some other topics, then we can follow different summarization strategies for those two classes of topics. Perhaps we can come up with class-based strategies to orient summarization systems towards different topical difficulty levels.

To probe these issues, we first define the notion of *summarizability*. The "summarizability" of a topic is defined as the mean of four ratings (from four systems) for that topic – each rating being an average across three annotators. This way, we obtain a single number for each of the 50 topics.

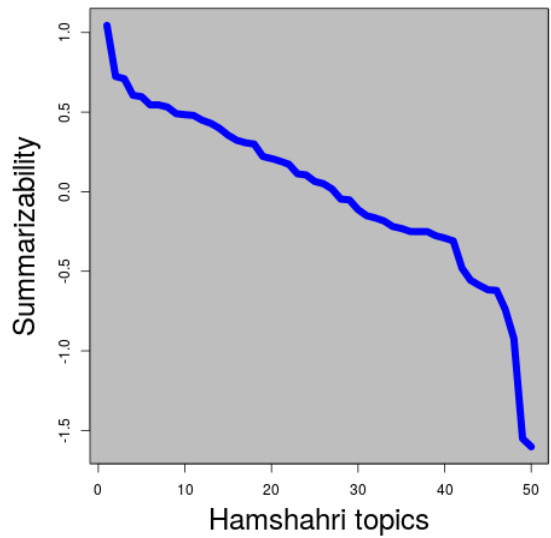

Figure 2: Summarizability of Hamshahri topics

Next, we sorted the summarizability values in descending order, and plotted them against topic ID (Figure 2). This plot shows a clear trend: a few topics are highly summarizable, a few are highly un-summarizable, and most fall in between. Hence, there are three classes of topics:

1. Highly summarizable topics (easiest to summarize)
2. Agnostic topics (in-between difficulty in terms of summarization)
3. Highly un-summarizable topics (hardest to summarize)

For Hamshahri, most of the topics turn out to be agnostic – which is a fairly broad range. Interestingly, this shows that Hamshahri – originally conceived as a dataset for Cross-Lingual Information Retrieval (CLIR) – is also a very good corpus for summarization research, with a mix of topics that touch different summarizability levels, and most of the topics either highly summarizable or agnostic. We show the topics along with their English description and summarizability values in Table 8 (sorted in descending order of summarizability).

| Topic | English Description | Summarizability |
|---|---|---|
| **607-AH** | Commemorations of Sadi Shirazi | 1.05 |
| **642-AH** | Shahr Theater Programs | 0.72 |
| **608-AH** | House Prices | 0.71 |
| **613-AH** | Children's Rights | 0.61 |
| **643-AH** | Earthquake Damage in Iran | 0.60 |
| **628-AH** | NATO vs. Yugoslavia in 1998 | 0.55 |
| **650-AH** | Fluctuations in Gas Imports | 0.55 |
| **629-AH** | Global Drought Predictions | 0.53 |
| **611-AH** | Information Technology and Employment | 0.49 |
| **612-AH** | Internet Users | 0.48 |
| **635-AH** | Iran in 1998 World Cup | 0.48 |
| **616-AH** | Hand-woven Carpet Exports | 0.45 |
| **619-AH** | Iranian Non-oil Exports | 0.43 |
| **622-AH** | Tehran Car Accidents | 0.40 |
| **644-AH** | Electronic Commerce | 0.35 |
| **617-AH** | Tourist Attractions | 0.32 |
| **609-AH** | Fruit Packing | 0.31 |
| **634-AH** | University Acceptance Limits | 0.30 |
| **623-AH** | North Iran Forestry Conservation | 0.22 |

| 615-AH | Remembrance of Dr Ali Shariati | 0.21 |
| --- | --- | --- |
| 618-AH | 7 July 1999 Protests | 0.19 |
| 631-AH | Relations between Iran and the United States | 0.17 |
| 639-AH | Bovine Spongiform Encephalopathy | 0.11 |
| 630-AH | Iranian Traditional Celebrations | 0.11 |
| 614-AH | E-commerce Congress | 0.06 |
| 625-AH | Places to Visit in Golestan | 0.05 |
| 647-AH | Buying Military Service Exemption | 0.02 |
| 632-AH | Olive Oil Benefits | -0.05 |
| 641-AH | Pollution in the Persian Gulf | -0.05 |
| 605-AH | Hatamikia's Films | -0.11 |
| 640-AH | Persian Rugs | -0.15 |
| 638-AH | Barriers for Investments in Iran | -0.16 |
| 610-AH | Benefits of Copyright Laws | -0.18 |
| 620-AH | Freight Transport by Rail | -0.22 |
| 602-AH | Heart Disease and Smoking | -0.23 |
| 606-AH | Youth Leisure in Summer | -0.25 |
| 626-AH | Women in Politics | -0.25 |
| 636-AH | Air Pollution | -0.25 |
| 646-AH | Applying to Study out of Iran | -0.28 |
| 648-AH | Attack on the Twin Towers | -0.29 |
| 637-AH | Tehran Air Pollution Sources | -0.31 |
| 621-AH | Television and Mental Health | -0.48 |
| 645-AH | 11 September and Air Travel | -0.56 |
| 649-AH | Khatami Government Oil Crisis | -0.59 |
| 624-AH | Films for the Fajr Festival | -0.62 |
| 633-AH | Daei's World Cup Goals | -0.62 |
| 603-AH | Gas Rationing in Iran | -0.74 |
| 627-AH | Nuclear Energy | -0.92 |
| 604-AH | Lung Cancer | -1.55 |
| 601-AH | US Attack on Iran | -1.60 |

Table 8: Summarizability of 50 topics

Note that about half of the topics have positive summarizability values, and the other half have negative. This is intuitive, because as we mentioned before, summarizability is coming from averaging several *z*-scores, and taken together, *z*-scores cancel each other out. Hence, we may expect to see (almost) equal number of positive and negative summarizability values.

Also of interest is the fact that the topics, when organized in this order, clearly show a trend: more concrete (and focused) topics tend to be highly summarizable, whereas more abstract (and diverse) topics – also, topics fraught with differing opinions, political controversy, etc – tend to be highly un-summarizable. Furthermore, local and national topics tend to be more summarizable than technical and international topics. A curious upshot of the above analysis is that most topics are from local and national news, and very few come from international. Even fewer relate to the United States, and contemporary friction arising from September 11 attacks. This shows a *social distinctiveness* of news topics. There is an inner core of densely connected and most popular *regional topics*, then there is an intermediate layer of loosely connected and somewhat less popular *national topics*, and finally there is the outermost layer (*periphery*) of sparsely connected and least popular *international topics*.

## 4.7. Question 7: Is There a Relationship Between Summarizability and Number of Documents?

We can reason that the *less* number of documents a topic has, the more focused (and marginal) it should be, and hence *easier* to summarize. In other words, there should be an inverse correlation between number of documents and summarizability. However, the following plot (Figure 3) – where X-axis is Hamshahri topics (sorted in descending order of summarizability), and Y-axis is the number of documents for each topic – shows that this is clearly not the case.

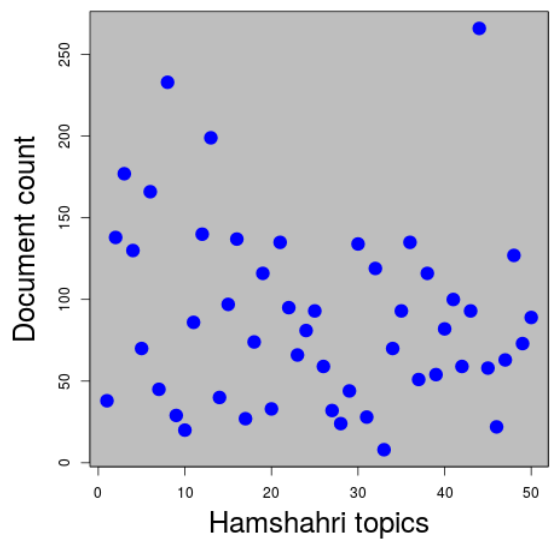

Figure 3: Document count of Hamshahri topics against summarizability

As seen from Figure 3, there is almost no correlation between the two variables. The Pearson correlation between summarizability and document count was found to be 0.09, which was not only too low, but also statistically indistinguishable from zero at 95% confidence level.

What this shows is that the issue of summarizability is complex, and needs to be investigated further. Perhaps there are other variables at play, e.g., document length, that have a greater bearing on how summarizability works, and hence can better explain the underlying process. Besides, it will be interesting to explore *classification* of Persian documents according to summarizability, because then we can *predict* – even before the actual summarization begins – whether the resulting summary will be any good. We leave this line of research to future work.

**5. Centrality-Based Summarization**

We already observed that centrality-based systems outperformed Parsumist-based systems in a *post hoc* analysis of multi-document summarization on the Hamshahri corpus. The next question we asked, is: *if we were given an annotated corpus of summarized Persian documents, which centrality measures would perform the best in single and multi-document summarization?* To answer this question, we resorted to the recently released Pasokh corpus of annotated Persian summaries (Behmadi Moghaddas et al., 2013). Note that we could have done this analysis post hoc, but that would have required our human annotators to go through hundreds of system-generated summaries manually – a clearly untenable approach. Instead, we used the popular ROUGE package (Lin, 2004) – in particular, its recent Java implementation called JROUGE (Ganesan et al., 2010) – to evaluate our centrality-based systems.

The rest of this section is organized as follows. In Section 5.1, we give an overview of centrality-based approaches, and discuss – using a cognitive science argument – why it makes sense to use them and why they perform so well in practice. Section 5.2 describes the systems we designed, and Section 5.3 details the evaluation strategy, including a description of Pasokh and ROUGE.

**5.1. Centrality-Based Methods**

It is well-known in distributional semantics that a word is known by the company it keeps (Firth, 1957). Words can have both local (and long-range) syntactic dependencies, and mostly global semantic dependencies. Hence, a *word network* (also variously known as a *collocation network* or a *collocation graph*; cf. (Lahiri, 2014)), where nodes are words and edges are word co-occurrence relationships, can succinctly and explicitly capture such dependencies. It has been observed (Mihalcea and Tarau, 2004) that in such networks, words with the highest *centrality* also happen to be the most important words (*keywords*) in the document. It has been similarly observed that if we construct the network on *sentences*, where nodes are sentences and edges are weighted (perhaps pruned, too) by sentence similarity, then the most important sentences in the document turn out to have highest centrality in the resulting network (Mihalcea and Tarau, 2004; Erkan and Radev, 2004).

These two observations started the field of graph-based summarization and keyword extraction (Mihalcea and Radev, 2011). Apart from the curious and interesting underpinnings as mentioned above, graph-based methods have the additional advantage that they are *unsupervised*, do not require any Natural Language Processing tools such as part-of-speech tagger, parser, or semantic role labeler, and still deliver acceptable (and in many cases, state-of-the-art) performance. The beauty and elegance of these approaches led to a flurry of research activities (cf. (Boudin, 2013; Lahiri et al., 2014)), and graph-based approaches are still being pursued by several research groups around the world.

It is interesting to note that the "word neighborhood" analogy holds for sentences too. It may initially come as a surprise how (or why) graph-based methods perform so well in practice, esp. when they take almost no help from existing Natural Language Processing tools. This is an important question, and the jury is still out. However, as Mihalcea and Tarau (2004) pointed out using a cognitive science argument, any network – be it words or sentences or people (social networks) – encodes a system of *endorsements* and

*recommendations*, where each edge denotes a form of endorsement. In essence, therefore, these networks are representing a *reputation model* where less important (thus, less reputed) words connect to more important (hence, more reputed) words, and less reputed sentences connect to more reputed sentences. If this model is taken as the *prima facie* proof of the success of graph-based approaches, then we can clearly see (and evidence suggests) that words and sentences with the most connections (i.e., highest centrality) are – in fact – the most important (reputed) words and sentences. Intriguingly enough, for summarization purposes, most important sentences are exactly what we need. Hence comes the general three-step approach to graph-based (or centrality-based) summarization:

1. Construct a network of sentences.
2. Rank the sentences by their centrality in the network.
3. Take top *k* of those sentences, and present them in temporal order.

Note that the above framework is very general, and can accommodate several types of networks and centrality measures. We under-specified the framework on purpose, so that we can play with different parameters. In practice, the concept of *centrality* has been studied in the social sciences for several decades (Borgatti, 2005). As Boldi and Vigna (2014) point out, there exist at least six different categories (*families*) of centrality indices that measure different aspects of a node in a network, as follows:

1. The node with largest degree
2. The node that is closest to other nodes
3. The node through which most shortest paths pass
4. The node with the largest number of incoming paths of length *k*, for every *k*
5. The node that maximizes the dominant eigenvector of the graph matrix
6. The node with the highest probability in the stationary distribution of the natural random walk on the graph

However, as Borgatti and Everett (2006) observed, most (if not all) of these indices assess a node's involvement in the *walk structure* of a network, and summarize a node's contribution to the *cohesiveness* of the network. Note that if language is represented as a network (of words or sentences), then the most important words/sentences should be the ones that contribute most highly to the cohesiveness of the document. So even from Borgatti and Everett's perspective, centrality-based approaches to summarization do make sense, and can be predicted to work well in practice.

**5.2. Centrality-Based Systems**

Following the general framework for graph-based summarization that we described in the previous section, it can be observed that there are essentially two free parameters: network type, and centrality measure. Network type can have important implications in a summarization system. Parsumist, for example, constructed a *complete graph* of sentences – then *pruned* the nodes that were deemed too similar to their neighbors. LexRank, on the other hand, went ahead with a complete graph representation and no pruning whatsoever. We, too, followed the same approach in Sections 3 and 4. Sentence *similarity* can also be an important aspect of network construction, because they allow us to *weight* the edges differently. While LexRank employs simple cosine similarity, TextRank used a modified form of *Dice Similarity*.

In this part of the study, we constructed sentence networks for both single and multi-document summarization. Sentences were extracted from the Pasokh corpus using the normalizer and tokenizer developed by Mojgan Seraji (Halácsy et al., 2007; Seraji, 2013; Seraji, 2015), in the same spirit as Section 3.1. Our networks are undirected and weighted complete graphs, where edge weights are given by cosine similarity between two sentences. Note that the network structure and choice of similarity measure will have an impact on the final result, but since our goal in this part of the study was to compare centrality measures, and not to tweak the network structure, we went ahead with the standard LexRank construction.

We experimented with three different vector representations of sentences – tf, tfidf, and binary (presence/absence of words). Vector elements are all unique words in the Pasokh corpus, and *idf* was computed on the Hamshahri corpus. Furthermore, we constructed sentence vectors in two ways – removing all stop words, and keeping them.[15] This gives rise to six types of sentence networks.

Once the sentence networks have been constructed, the next step is to select the most *central* sentences in the network. Note that the use of centrality measures in document summarization is not new, but it is new in Persian. We used the following seven centrality measures:[16]

1. **Strength:** sum of the weights of the edges incident to a node (also called "weighted degree").

---

[15]Stop words came from the Hamshahri collection, available at
http://ece.ut.ac.ir/dbrg/hamshahri/files/HAM1/persian.stop.txt.
[16]Centrality was computed using the *igraph* package (Csardi and Nepusz, 2006).

2. **Clustering Coefficient:** density of edges among the immediate neighbors of a node (Watts and Strogatz, 1998).
3. **Structural Diversity Index:** normalized *entropy* of the weights of the edges incident to a node (Eagle et al., 2010).
4. **PageRank:** importance of a node based on how many important nodes it is connected to (Page et al., 1998).
5. **Betweenness:** fraction of shortest paths that pass through a node, summed over all node pairs (Anthonisse, 1971; Brandes, 2001).
6. **Closeness:** reciprocal of the sum of distances of all nodes to a node (Bavelas, 1950).
7. **Eigenvector Centrality:** element of the first eigenvector of a graph adjacency matrix corresponding to a node (Bonacich, 1987).

The above seven centrality measures touch all six different centrality families identified by Boldi and Vigna (2014), hence they may be deemed comprehensive for the purposes of our study. Note further that all the above measures except Structural Diversity Index above are *true centrality indices*, hence sentences need to be sorted in their *descending order*. For the Structural Diversity Index, however, we want sentences to have *minimum* possible values, so they need to be sorted in the *ascending order*. Lastly, we used *strength* instead of degree, because our networks are complete graphs, hence all nodes have the same degree.

The seven centrality measures we used gave us 42 graph-based systems in the end (six types of graphs, seven centrality measures on each of them). In the evaluation, we will see that most systems behave rather similarly, and differences emerge from centrality measures rather than narrow, document-specific optimizations. In particular, strength, PageRank, eigenvector centrality and Structural Diversity Index perform the best in terms of ROUGE score.

### 5.3. Evaluation of Centrality-Based Systems

As mentioned in Section 3.6, evaluation is the Achilles' heel for most summarization systems, which is exacerbated by the fact that oftentimes, no human-annotated gold standard dataset is available on which competing systems can be evaluated. We were rather fortunate in this regard, because recently such a gold standard corpus – Pasokh – has been released by a team of researchers that specifically looks into Persian single and multi-document summarization (Behmadi Moghaddas et al., 2013).

Pasokh is a corpus of Persian news articles annotated with human-generated summaries – both *extractive* and *abstractive* – for single-document as well as multi-document summarization. It took over 2000 man-hours of work to construct Pasokh. The meaning of the word "Pasokh" is "answer"; it is a contraction of the original Persian: *Peykare-ye (A)estândârd-e Sâmânehâ-ye (O is added for ease of pronunciation) KHolâse-sâz*. The news content of Pasokh comes from seven popular Persian news agencies (including Hamshahri), and six genres of documents (Economic, Cultural, Social, Political, Sports, and Scientific). Pasokh has 100 documents in the single-document summarization section. Each document has five extractive and five abstractive summaries that come from five different human annotators. In total, this gives us 500 extractive and 500 abstractive summaries to work with.

The multi-document section of Pasokh comprises 50 *news topics*, 20 documents per topic. The same seven news agencies were used. Five extractive and five abstractive summaries were created for *each topic* – by five different human annotators – at a 30% compression ratio. So in this section, we obtain 250 extractive and 250 abstractive summaries.

Ten male and ten female undergraduate students constructed the corpus. They were trained to represent the *central content* of each document/topic in the summary, while at the same time avoiding repetition and redundancy; according with the key points of the original text; maintaining coherence and readability, particularly in the abstractive summaries; maintaining cohesion; and not exceeding the set compression ratio (30%). The large size and complexity of the corpus forced its creators to design a separate software interface called *Kholâse-yâr* (roughly translates as "summarization aid") to help the human annotators. The annotators had to supply reasons for why they chose a particular piece of text to be included in the summary. Thus, Pasokh turned out to be a very exhaustive, complete, and holistic corpus, and a great exercise in human annotation in Persian language. For single documents, Pasokh summaries are mostly three to seven sentences long.

We used the extractive portion of Pasokh, and left the abstractive part for future work. Note that the extractive summaries in Pasokh do not always obey strict extraction rules. For example, many of them extract phrases and clauses rather than full sentences, and then "glue" together components in an abstractive fashion. Hence, the extractive summaries are not purely extractive – they oftentimes contain abstractive components. This observation makes standard evaluation using precision, recall, and F-score (at sentence level) difficult on the Pasokh extractive set. We used the ROUGE score (Lin, 2004) instead.

The word "ROUGE" stands for *Recall-Oriented Understudy for Gisting Evaluation*. ROUGE essentially measures *overlap* between system-generated summaries (also called "peer summaries") with human-generated summaries (also called "model summaries"). Note that recall is more important in the summarization setting than precision, because we would like to have *as many* important units of information as possible in the final summary, rather than only the top few most salient ones. There are four different ROUGE measures: ROUGE-N, ROUGE-L, ROUGE-W, and ROUGE-S. ROUGE-N is based on n-gram co-occurrence statistics, ROUGE-L is based on *longest common subsequences* (LCS) at the summary level, ROUGE-W represents *weighted* longest common subsequences, and ROUGE-S involves skip-bigram co-occurrence statistics.

As one can notice, the ROUGE scores become progressively more complex from ROUGE-N to ROUGE-S. In practice, most researchers use ROUGE-N, because it is simple to understand, and gives good results. We, too, used ROUGE-N. The formula for ROUGE-N is as follows:

$$ROUGE_N = \frac{\sum_{S \in \{ModelSummaries\}} \sum_{gram_n \in S} Count_{match}(gram_n)}{\sum_{S \in \{ModelSummaries\}} \sum_{gram_n \in S} Count(gram_n)} \quad (6)$$

where *n* is the length of the n-gram, *S* is a particular human-generated summary, $gram_n$ is a particular n-gram, and $Count_{match}(gram_n)$ is the maximum number of n-grams co-occurring in a system-generated summary and a set of human-generated summaries.

Note that ROUGE talks about *a set of human-generated summaries* – not a *single* human-generated summary. This is typical in summarization research. Owing to the high variability between human annotators, we would like to generate as many model summaries as we can, and then evaluate the peer summaries against *all of them*. This is a time-tested way to combat variability in Natural Language Processing tasks such as keyword extraction and summarization. Note further that the standard ROUGE-N formula above can be adapted to make way for a precision-oriented measure, which in turn gives rise to an F-score-style measure. In our evaluation, we used all three (precision, recall, and F-score). F-score combines precision and recall in a theoretically sound way, but as we will see in Section 6, precision and recall showed their own interesting trends.

Interestingly, it would not have been possible to use ROUGE in Sections 3 and 4 (i.e., on the Hamshahri corpus), because there we did not have human-annotated gold-standard summaries (*model summaries*). A particular problem with ROUGE is that its vanilla implementation in Perl does not handle Unicode characters. Hence, we resorted to the recently released JROUGE package written in Java (Ganesan et al., 2010). We used three baselines to compare our centrality-based systems against:

1. Random *k* sentences extracted from the document.
2. First *k* sentences extracted from the document.
3. Last *k* sentences extracted from the document.

While random baseline is a common staple in almost all Natural Language Processing evaluation tasks, the first and last *k* sentences baselines are special in summarization. They derive their importance from the fact that in a document (esp. news articles like Pasokh and Hamshahri), first *k* sentences often contain the most important and/or the most central pieces of information. Hence, a summary consisting of first *k* sentences often constitutes a very strong and very effective baseline in practice. On the other hand, the last *k* sentences in a document often include concluding remarks and thoughts on future directions or courses of events. Hence, those sentences could turn out to be important in terms of summarization. We therefore opted for all three baselines in this study. We used the 42 centrality-based systems, and the above three baseline systems at compression ratios of 25%, 50%, 75%, and 100%, respectively.

## 6. Results on the Pasokh Corpus

As mentioned in Section 5.3, we used ROUGE-N for evaluation. We experimented with three values of *N*: 1, 2, and 3. These correspond to ROUGE on unigrams, ROUGE on bigrams, and ROUGE on trigrams, respectively. They are standard metrics in summarization literature. What we observed is that qualitatively, ROUGE-1, ROUGE-2 and ROUGE-3 are similar to each other. Hence, we report ROUGE-3 only. Quantitatively, ROUGE-1 is higher than ROUGE-2, which in turn is higher than ROUGE-3. This is expected, because between any two pieces of text, the number of common unigrams should be at least as many as the number of common bigrams, and the number of common bigrams should be at least as many as the number of common trigrams. This makes ROUGE-3 the most pessimistic (*conservative*) of all three ROUGE-N scores we tried.

Furthermore, note that we used three vectorial representations: binary, tf, and tfidf. We observed that they had very little difference among themselves – qualitatively and quantitatively. Hence, we only report tfidf. In the following, we encode our centrality-based systems and baseline systems using a three-character code, as follows:

- **Bet:** Betweenness centrality
- **Clo:** Closeness centrality
- **Clu:** Clustering Coefficient
- **Div:** Structural Diversity Index

- **Eig:** Eigenvector Centrality
- **Pag:** PageRank
- **Str:** Strength
- **Fir:** First *k* sentences baseline
- **Las:** Last *k* sentences baseline
- **Ran:** Random *k* sentences baseline

Lastly, in the following stacked bar diagrams, the lowest level corresponds to 25% compression ratio, second lowest level 50% compression ratio, third lowest (i.e., second highest) level 75% compression ratio, and the highest level 100% compression ratio. They are denoted by four different colors for ease of visualization.

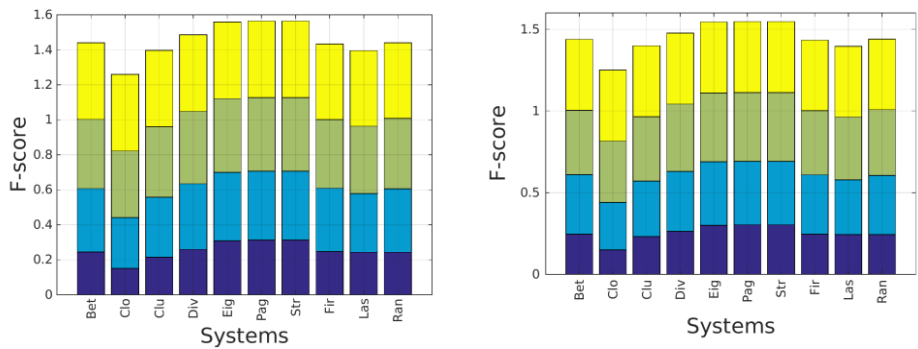

Figure 4: ROUGE-3 F-score on *single-document summarization*; without stop words on left, with stop words on right.

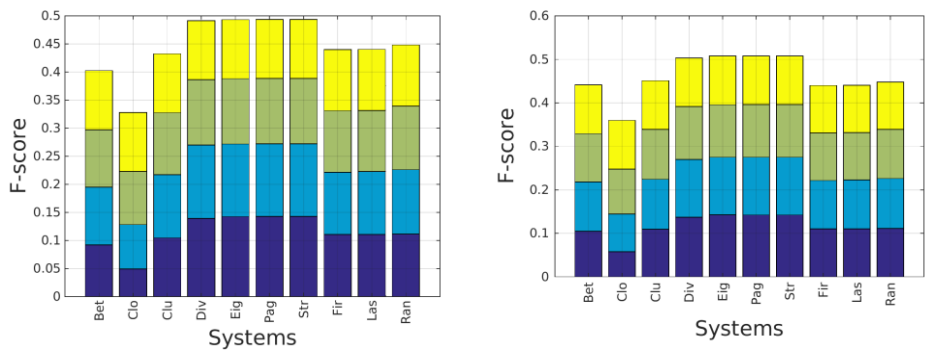

Figure 5: ROUGE-3 F-score on *multi-document summarization*; without stop words on left, with stop words on right.

As seen from Figures 4 and 5, with stop words and without stop words systems perform quite similarly in terms of F-score across all compression ratios. Overall, best F-scores are realized by strength, PageRank, eigenvector centrality, and Structural Diversity Index. It is intriguing to see that PageRank is not the only best-performing measure. Presence of Structural Diversity Index among the best performers is another intriguing finding, and should be researched in more depth. Interestingly, all three baselines perform quite well, esp. compared to betweenness, closeness, and clustering coefficient. Compression-ratio-wise, 50% and up seems good enough for single-document summarization. However, we should keep in mind that for long documents, 50% may be too high. For multi-document summarization, best results come at 25% compression ratio. This is intuitive, because for multiple documents, 25% is already large enough to contain most (if not all) units of importance.

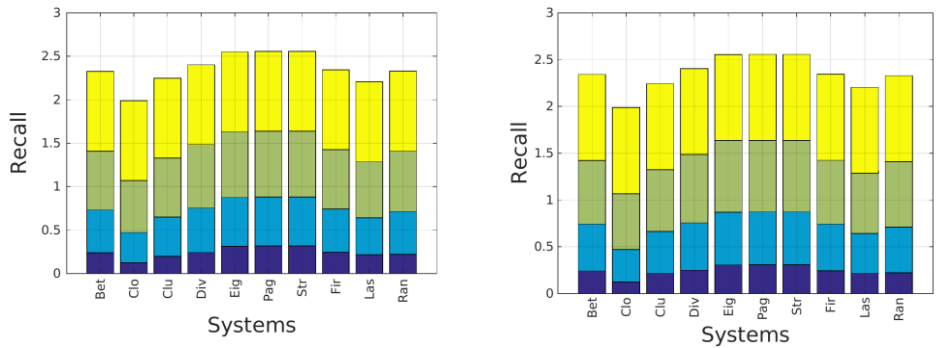

Figure 6: ROUGE-3 recall on *single-document summarization*; without stop words on left, with stop words on right.

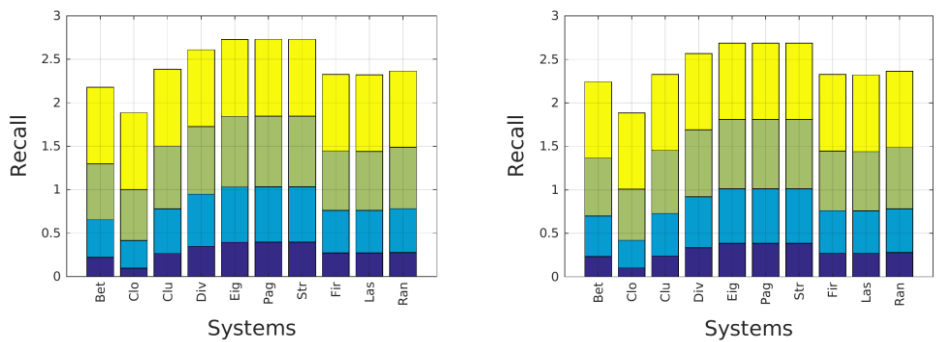

Figure 7: ROUGE-3 recall on *multi-document summarization*; without stop words on left, with stop words on right.

We report ROUGE-3 *recall* in Figures 6 and 7. Note from the discussion in Section 5.3 that recall is in fact the most important performance measure in summarization. As seen from Figures 6 and 7, recall increases with increasing compression ratio, and the best values are obtained at the maximum compression ratio of 100% (i.e., no compression). This makes intuitive sense, because as more and more units of information are included in the summary, it becomes more and more probable to find a particular relevant piece of information in them. Hence, recall goes up.

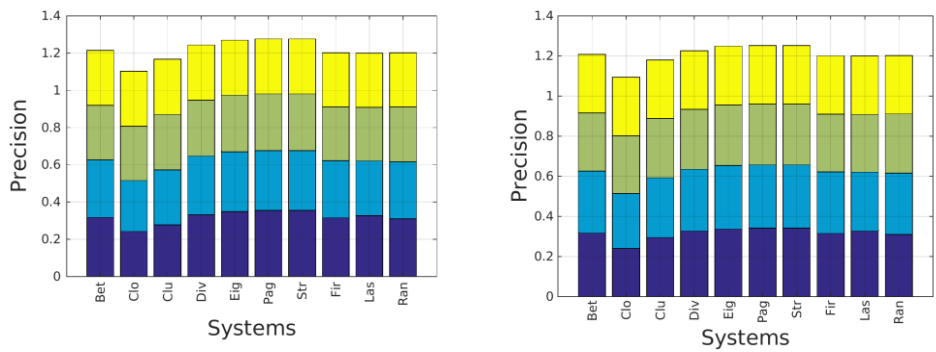

Figure 8: ROUGE-3 precision on *single-document summarization*; without stop words on left, with stop words on right.

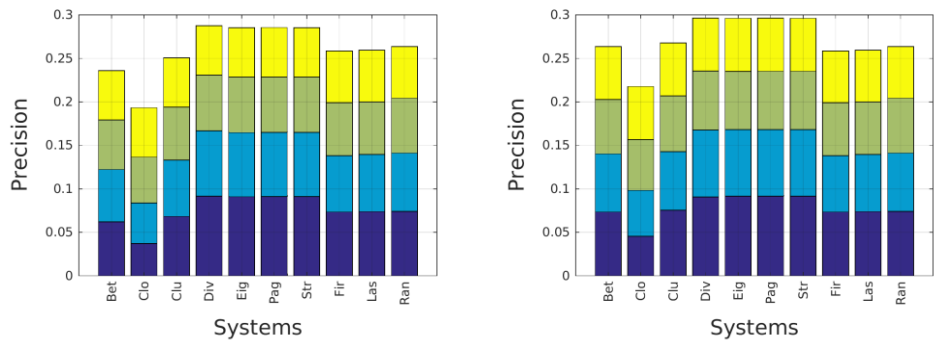

Figure 9: ROUGE-3 precision on *multi-document summarization*; without stop words on left, with stop words on right.

The opposite happens with ROUGE-3 *precision* (Figures 8 and 9). As compression ratio goes *up*, precision goes *down*, because we are including many more irrelevant pieces of information in the summary than relevant ones. This dichotomy of precision and recall is a standard phenomenon in Information Retrieval (IR), but it was interesting to observe the same dichotomy hold true in the domain of summarization and ROUGE – especially in Persian.

**7. Conclusion and Future Work**

We have described three classes of novel unsupervised systems for single and multi-document summarization in Persian – one based on Parsumist, a popular Persian summarizer, and two based on sentence networks and centrality measures. Experimental results indicate that graph-based methods perform better than Parsumist-based methods, and there are some centrality measures (strength, PageRank, eigenvector centrality, and Structural Diversity Index) that perform better than others. Detailed human evaluation (*post hoc*), as well as evaluation using ROUGE score have been reported.

There are several limitations of our study, all of which can be improved upon in future work:

- Our study does not consider the problem of *supervised* summarization, where a machine learning system is trained to identify sentences and/or other linguistic units for inclusion in the summary. Existing literature suggests that supervised systems perform better than unsupervised ones.
- We have not been able to compare our systems with state-of-the-art Persian summarizers, because their implementation is not publicly available. It will be an interesting and revealing exercise to follow up on this lead.
- Parameter tuning and optimization: we performed minimal parameter tuning. There are many opportunities to do so, esp. in Parsumist, and/or graph-based systems. Future work should look into parameter tuning with more rigor and depth.
- We did not attempt abstractive summarization in this work, because it involves non-trivial amount of syntactic judgment calls. Since we did not have a Persian linguist in our team, such judgment calls were deemed arbitrary and hard to justify. Future work should look into this very interesting and exciting line of research.
- We did not use syntactic and semantic features. Dependency treebanks have been developed and are currently available for Persian (Rasooli et al., 2013; Seraji, 2015). It will be a very promising and interesting research direction to integrate such resources with extractive and/or abstractive summarization in Persian.

Overall, we believe that we have presented a convincing study in Persian document summarization with existing resources, and hope that our work will spark interest in the still nascent but growing area of research in document summarization in Persian.